\pdfoutput=1
\documentclass[11pt]{article}

\usepackage[]{EMNLP2022}

\usepackage{times}
\usepackage{latexsym}

\usepackage[T1]{fontenc}
\usepackage[utf8]{inputenc}

\usepackage{microtype}

\usepackage{inconsolata}

\usepackage{microtype}
\usepackage{graphicx}
\usepackage{subfigure}
\usepackage{amsmath}
\usepackage{amsfonts}
\usepackage{multirow}
\usepackage{color}
\definecolor{Red}{rgb}{1,0,0}
\definecolor{Green}{RGB}{0,153,1}
\definecolor{Blue}{rgb}{0,0,1} 
\usepackage{tablefootnote}

\title{Pair-Based Joint Encoding with Relational Graph Convolutional Networks for Emotion-Cause Pair Extraction}

\author{Junlong Liu, Xichen Shang, Qianli Ma\footnotemark[1] \\
  School of Computer Science and Engineering, \\ 
  South China University of Technology, Guangzhou, China \\
  \texttt{junlongliucs@foxmail.com} \\ \texttt{qianlima@scut.edu.cn\footnotemark[1]}}

\begin{document}
\maketitle

\renewcommand{\thefootnote}{\fnsymbol{footnote}}
\footnotetext[1]{Corresponding author}

\renewcommand{\thefootnote}{\arabic{footnote}}

\begin{abstract}
Emotion-cause pair extraction (ECPE) aims to extract emotion clauses and corresponding cause clauses, which have recently received growing attention. Previous methods sequentially encode features with a specified order. They first encode the emotion and cause features for clause extraction and then combine them for pair extraction. This lead to an imbalance in inter-task feature interaction where features extracted later have no direct contact with the former. To address this issue, we propose a novel \textbf{P}air-\textbf{B}ased \textbf{J}oint \textbf{E}ncoding (\textbf{PBJE}) network, which generates pairs and clauses features simultaneously in a joint feature encoding manner to model the causal relationship in clauses. PBJE can balance the information flow among emotion clauses, cause clauses and pairs. From a multi-relational perspective, we construct a heterogeneous undirected graph and apply the Relational Graph Convolutional Network (RGCN) to capture the various relationship between clauses and the relationship between pairs and clauses. Experimental results show that PBJE achieves state-of-the-art performance on the Chinese benchmark corpus.\footnote{Our codes are publicly available at \url{https://github.com/tutuDoki/PBJE-ECPE}}
\end{abstract}

\section{Introduction}
\label{sec:introduction}
\begin{figure}[t!]
    \centering
    \includegraphics[width=\linewidth]{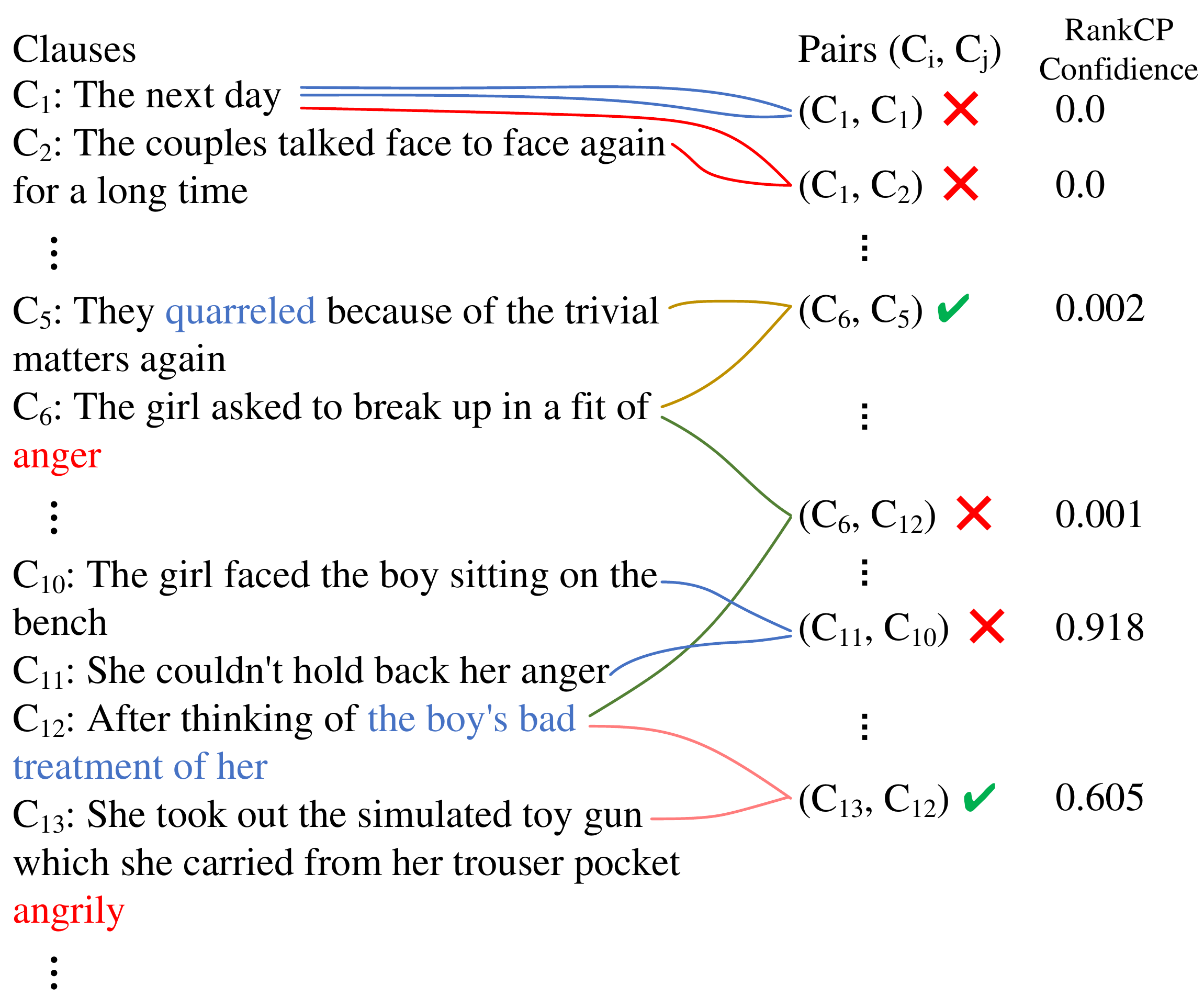}
    \caption{An example document from the ECPE corpus where ${c}_i$ represents the emotion clause and ${c}_j$ represents the cause clause in pair. The words in \textcolor{Red}{red} are the keywords about emotion and the words in \textcolor{Blue}{blue} are about cause.
    The emotion clause $c_6$ and $c_{12}$ can not be a pair for lack of causal relationship. 
    We translate it from Chinese into English for ease of reading.}
    \label{fig:example}
\end{figure}

Emotion cause extraction (ECE) is a kind of emotion analysis task which is first proposed by \citet{lee-etal-2010-text} and has developed for a long time \citep{gui-etal-2017-question, li-etal-2018-co, LI2019205, hu-etal-2021-bidirectional-hierarchical}. ECE extracts the cause for the input document and certain emotion labels. However, emotions in the documents need to be annotated in advance, which requires manual input and takes lots of time \citep{xia-ding-2019-emotion}. Hence, \citet{xia-ding-2019-emotion} proposes a new task called emotion-cause pair extraction (ECPE). Given a document as the input, ECPE extracts the clauses which express emotions and their corresponding clauses which express causes (as shown in Figure~\ref{fig:example}). Intuitively, ECPE is much more challenging because the clauses classification task and the pairs matching task need to be completed simultaneously.

For ECPE, \citet{xia-ding-2019-emotion} first proposes a two-stage method. However, the two-stage method may cause the problem of error propagation. To solve this problem, the previous work uses end-to-end methods \citep{ding-etal-2020-end,chen-etal-2020-end,singh-etal-2021-end}. Most of them use sequential encoding, in which their task-specific features are learned sequentially in a predefined order. Specifically, following \citet{wei-etal-2020-effective}, ECPE contains two auxiliary tasks, which are emotion clause extraction (EE) and cause clause extraction (CE). The previous work first separately models the clauses for EE and CE. Then they use the clause representations for emotion and cause clauses to model the pairs for ECPE. 

However, the sequential encoding only considers the intra-relationship within pairs or clauses while ignoring the inter-relationship between them. In the sequential encoding, the information can only flow from emotion/cause clause encoder to pair encoder but not vice versa, resulting in the exposure of different amounts of information to pair encoder and clause encoders \citep{yan-etal-2021-partition, wang2022towards}. In this way, if the emotion/cause clause encoder makes incorrect predictions, it will severely misguide the predictions of pair. For example, in the Figure~\ref{fig:example}, the previous model RankCP \citep{wei-etal-2020-effective} wrongly extracts the $c_{11}$ as an emotion clause and the $c_{10}$ as a cause clause with high degree of confidence (about 0.98 for $c_{11}$ and 0.94 for $c_{10}$). Finally, the pair encoder of RankCP extracts the couple $(c_{11}, c_{10})$ as a candidate pair, which is an incorrect answer. The imbalanced information flow (i.e., from clauses to pairs) might have induced this error.
% And adjusting this unreasonable information flow can help correct these typical wrong cases. 

On the contrary, the joint encoding manner is getting more attention in multi-task learning \citep{lai-etal-2021-joint,wang-etal-2020-tplinker,fu-etal-2019-graphrel,wang-lu-2020-two}. It can not only balance the information flow among emotion clauses, cause clauses and pairs to deal with the problems of sequential encoding we mentioned above but also take into account the causal relationship \citep{chen-etal-2020-conditional} between clauses in ECPE.
Since the joint encoding can make the emotion/cause clause encoder and pair encoder interact with each other.
During the process of encoding, the clause encoders can pay more attention to whether a clause is suitable for pairs rather than only focus on the information about emotion or cause. The causal relationship is a decisive factor in judging whether emotions and causes match. For example, in Figure~\ref{fig:example}, ${c}_6$ and ${c}_{13}$ both express anger, and ${c}_{12}$ is cause clause. However, $(c_{13}, c_{12})$ is a pair and $(c_6, c_{12})$ is not. If we separately model the pairs and clauses, the lack of relationship information between these two clauses will increase the difficulty for the model to judge this situation.

Given the above mentioned situation, we propose a novel \textbf{P}air-\textbf{B}ased \textbf{J}oint \textbf{E}ncoding (\textbf{PBJE}) method, which simultaneously generates pairs and clauses features in a joint feature encoding manner. Specifically, we model the inter-relationship between pairs and clauses, in which a pair only interacts with the corresponding clauses. It helps pairs learn representations and model the causal relationship from clauses.
% Besides, it can prevent interference from irrelevant information. 
Meanwhile, the key information about emotion and cause clauses is different. Therefore, different features should be extracted from these two clauses. Considering these multiplex relationships, we construct a heterogeneous undirected graph and apply Relational Graph Convolutional Networks (RGCN) \citep{schlichtkrull2018modeling} on it, which includes four kinds of nodes and five kinds of edges, utilizing different approaches to connect the nodes. Thus, it can make the information flow between emotion clauses, between emotion clauses and pairs, etc., more efficient.

We summarize our contributions as follows: (1)We propose a novel method called PBJE to jointly encode the clauses and pairs for ECPE, helping the pairs learn the causal relationship between the two clauses during the encoding process. (2)We propose a RGCN framework to model the multiplex relationship between pairs and clauses. Different edges in the RGCN help the pairs or clauses extract more targeted information, improving the efficiency of the information flow. (3)Experiments on ECPE benchmark corpus demonstrate that PBJE is state-of-the-art. 
% Furthermore, some other experiments are performed to verify the effectiveness of PBJE.
% \begin{itemize}
%     \item  We propose a novel method called PBJE to jointly encode the clauses and pairs for ECPE, helping the pairs learn the causal relationship between the two clauses during the encoding process.
%     \item  We propose a RGCN framework to model the multiplex relationship between pairs and clauses. Different edges in the RGCN help the pairs or clauses extract more targeted information, improving the efficiency of the information flow.
%     \item  Experiments on ECPE benchmark corpus demonstrate that PBJE is state-of-the-art. Furthermore, some other experiments are performed to verify the effectiveness of PBJE.
% \end{itemize}

\section{Related Work}

\subsection{Sequential Encoding}
Most of the previous work uses sequential encoding to solve ECPE, including the pipeline and unified framework. Specifically, \citet{xia-ding-2019-emotion} proposes ECPE task and two auxiliary tasks (EE and CE). It uses a two-stage method that first extracts the emotion and cause clauses and then matches them as pairs using Cartesian product for prediction. To address the error propagation problem, \citet{wei-etal-2020-effective} proposes a unified framework that uses Graph Convolution Networks to encode the emotion and cause clauses in the same representations. However, it does not model the pairs, leading to a lack of contextual information in pairs. Furthermore, \citet{ding-etal-2020-ecpe, ding-etal-2020-end} and \citet{chen-etal-2020-end} build encoders for pairs and clauses separately, which model clauses and then concatenate them as pairs. Considering the symmetric relation between emotion clauses and cause clauses, \citet{cheng-etal-2020-symmetric} uses a local search method for the clauses which are predicted as emotion clauses or cause clauses.

However, these typical sequential encoding models encode the features in a predefined order, which leads to the imbalance of the inter-task feature interaction. Since the interaction between clauses and pairs is unidirectional, and the features in pairs can not flow to clauses.

\subsection{Implicit Joint Encoding}
On the other hand, some work solve ECPE with the implicit joint encoding, such as the sequence labeling methods. We call them "implicit" because these methods joint encode the clauses and pairs, but they do not have the apparent pair features in the model. For example, \citet{yuan-etal-2020-emotion} designs a novel cause-pivoted tagging theme. This theme first predicts if a clause is a cause clause, and then finds the corresponding emotion clause among its neighbors using the relative position. In addition, \citet{9457144} propose a tag distribution refinement method based on the cause-pivoted sequence labeling, which can leverage the correlations between different tasks (i.e., ECPE, EE, and CE) explicitly and exploit information interaction. Due to the drawback of the cause-pivoted, which can not perfectly extract the cause clauses with multiple emotion clauses, \citet{chen-etal-2020-unified} designs a more fine-grained tagging scheme that combines emotion tagging and cause tagging with emotion labels separately. However, it still can not handle the situation in which a document has multiple pairs with the same type of emotions. Given this, \citet{9511845} designs a special set of unified labels based on the sequence to sequence model.

Nonetheless, these implicit joint encoding methods based on sequence labeling lack the explicit interaction between clauses and pairs compared with our explicit joint encoding manner. This means that much causal relationship information is ignored in these methods.

% To address the issues we mentioned above, we propose an explicit joint model based on the Relational Graph Convolutional Network, which can balance the information between clauses and pairs and fully consider the causal relationship between emotion clauses and cause clauses.

\begin{figure*}[t!]
    \centering
    \includegraphics[width=\linewidth]{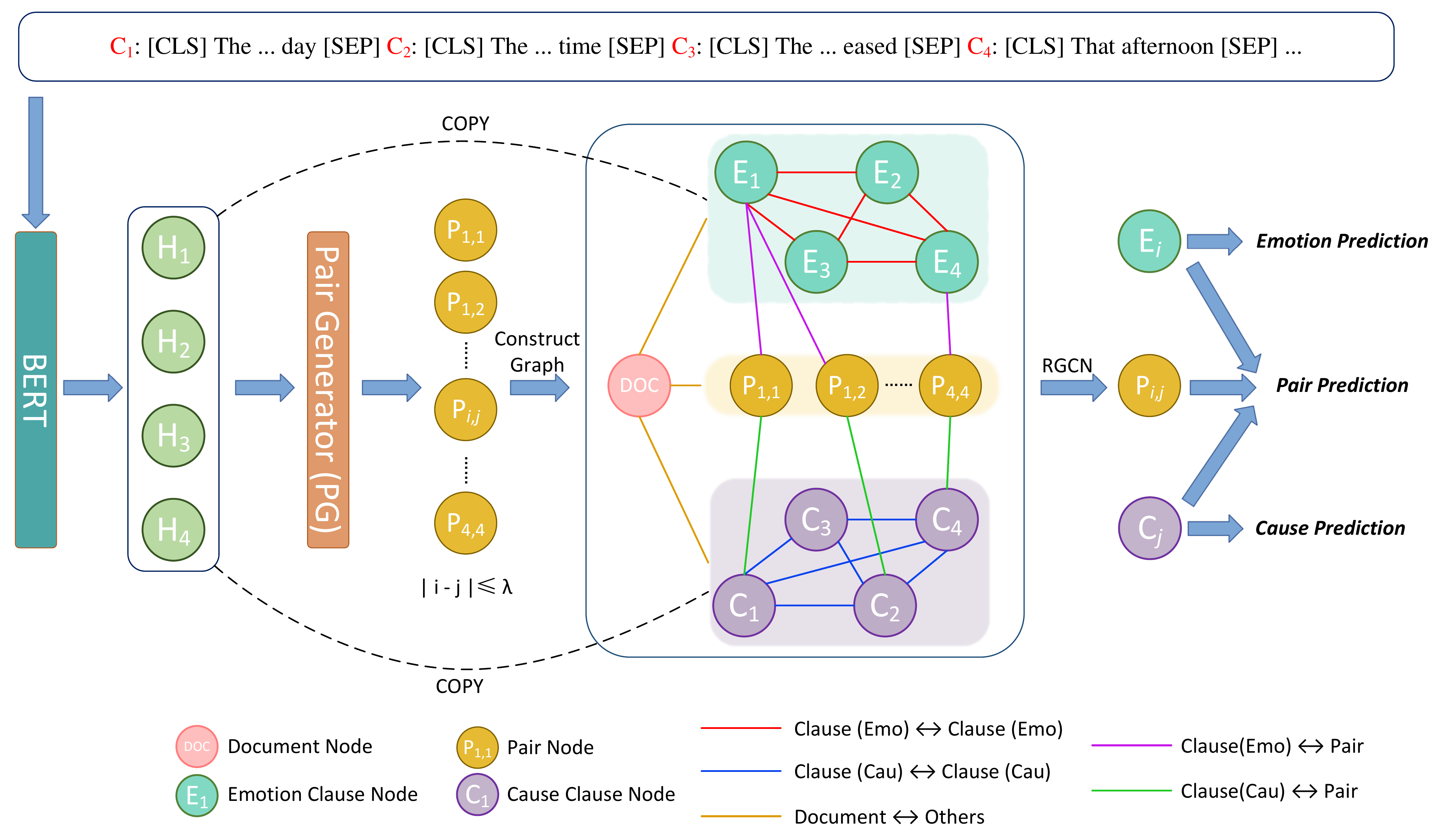}
    \caption{The overall architecture of the PBJE. First, the clauses are input into the pre-trained BERT to get the representations. Then, the representations of the pair are obtained by the pair generator. Next, we construct a heterogeneous undirected graph with the emotion clause nodes, cause clause nodes, pair nodes, and a document node. Finally, after applying RGCN, we use the last layer's representations of the node for predictions.}
    \label{fig:model}
\end{figure*}

\section{Task Definition}
Given a document $D=(c_1, c_2, \ldots, c_{N})$ of $N$ clauses and the $i$-th clauses $c_i=(w_1^i, w_2^i, \ldots, w_M^i)$ of $M$ words, ECPE task aims to extract all the emotion-cause pairs in $D$:
    \begin{equation}
        P=\left\{\ldots,(c_i, c_j),\ldots\right\} \quad (1 \leq i, j \leq N)
    \end{equation}
where $c_i$ and $c_j$ represent the emotion clause and corresponding cause clause in pairs.

Meanwhile, ECPE has two auxiliary tasks which are emotion clauses extraction (EE) and cause clauses extraction (CE). A clause $c_i$ is emotion clause if any pair $(c_i, c_j)$ is established. It can be defined as follow:
    \begin{equation}
        y_i^{emo} = 
        \begin{cases}
            1, &if \quad {\exists} c_j \in D, (c_i, c_j) \in P
            \cr 0, & otherwise
        \end{cases}
    \end{equation}
where $y_i^{emo} = 1$ means $c_i$ is an emotion clause. The extraction of cause clauses is the same as emotion clauses.

\section{Proposed Model}
In this section, we mainly describe our method, which encodes the pairs and clauses simultaneously and models the causal relationship from clauses in Relational Graph Convolutional Network (RGCN). The structure of PBJE is shown in Figure~\ref{fig:model}.

\subsection{Pair Generator}
Following \citet{wei-etal-2020-effective}, given a document $D = (c_1, c_2, \ldots, c_N)$ consisting of $N$ clauses, we feed $D$ into pre-trained BERT \citep{devlin-etal-2019-bert}. Specifically, we add a token $[\text{CLS}]$ at the beginning and a token $[\text{SEP}]$ at the end for each clause and concatenate all clauses together as input. Finally, we use the average pooling of the representations of tokens except for the $[\text{CLS}]$ and $[\text{SEP}]$ in each clause as the representations of clauses.
Hence, the document with $N$ clauses can be represented as:
    \begin{equation}
        H = \left\{ h_1,  h_2, \ldots, h_N\right\}
    \end{equation}
where $h_i \in \mathbb{R}^{d}$ and $d$ is the hidden size of BERT.

To obtain the representations of pairs, we employ the Pair Generator (PG). Specifically, we concatenate the corresponding two clauses and project them with a learnable relative position embedding:
    \begin{equation}
        p_{ij} = W_p [h_i, h_j] + b_p + r_{i-j}
    \end{equation}
where $p_{ij} \in \mathbb{R}^d$ represents the pair using $c_i$ as an emotion clause and $c_j$ as a cause clause, $W_p \in \mathbb{R}^{d \times 2d}$ and $b_p \in \mathbb{R}^d$ are learnable parameters, $r_{i-j} \in \mathbb{R}^d$ is the relative position embedding, and $[,]$ denotes the concatenating operation. In addition, following \citet{wei-etal-2020-effective}, we set a hyperparameter $\lambda$ as the local window $(|i - j| \le \lambda)$ to limit the number of pairs.

\subsection{Pair-Based Joint Encoder}
To balance the interaction between pairs and clauses and capture the causal relationship in pairs, we construct a heterogeneous undirected graph. It can deal with the various relationship between pairs and clauses as well as the relationship between clauses efficiently.

The graph has four kinds of nodes: emotion clause nodes, cause clause nodes, pair nodes, and a document node. The emotion information and cause information in a clause are contained in different words. Hence, we separately use two kinds of nodes to represent the emotion clause and the cause clause. In addition, to directly interact with the clauses and capture the causal relationship between the corresponding emotion clause and cause clause, pair nodes are added to the graph. The simultaneous encoding of clauses and pairs balances the information flow between them. Meanwhile, we add a document node to the graph, which can provide some global information (e.g., topics) for the other nodes and interact with others like a pivot.

Moreover, there are mainly five kinds of inter-node edges in our graph:
\begin{itemize}
    \item \textbf{Clause-Clause Edge:} There are two kinds of clause edge in our graph, including the Clause(Emotion)-Clause(Emotion) and Clause(Cause)-Clause(Cause). All emotion/cause clause nodes are fully connected with their own edge. These two edges can help each emotion/cause clause node interact with other emotion/cause nodes to access contextual information.
    \item \textbf{Clause-Pair Edge:} There are also two kinds of Clause-Pair edge in our graph, including the Clause(Emotion)-Pair and Clause(Cause)-Pair. All pair nodes are connected to their corresponding emotion clause nodes and cause clause nodes with these two kinds of edge. They are the primary way for pairs and clauses to interact with each other and help the emotion and cause nodes to transmit causal relationship to the pair nodes. Besides, the emotion and cause nodes can interact through the pair nodes and these two edges.
    \item \textbf{Document-Others Edge:} The document node is connected to all other nodes with this edge. It can transmit the global information in the document to other nodes and help other nodes ignore the noise from irrelevant nodes.
    % We do not use different edges to separately connect the clause nodes and pair nodes since we argue that the global information is the same for all kinds of nodes. Too many learnable parameters will make the model hard to train.
\end{itemize}

Furthermore, each type of node has a kind of self-loop edge, which can help each node to maintain its feature in the process of interaction.

With jointly encoding the pairs and clauses at the same level in the graph, the model can solve the problems in sequential encoding and balance the information between pairs and clauses.

Next, the Relational Graph Convolutional Network (RGCN) \citep{schlichtkrull2018modeling} is applied on our heterogeneous undirected graph to aggregate the features from neighbors of each node. First, we use the representations of clause to initialize each emotion and cause clause node:
    \begin{gather}
        H^{(0)}_E = H, \, H^{(0)}_C = H
    \end{gather}
where $H^{(0)}_E$ is the representations of emotion clause nodes and $H^{(0)}_C$ is the representations of cause clause nodes. Then, we use the representations of pairs to initialize the pair nodes:
    \begin{equation}
        H^{(0)}_P = \left\{p_{11}, p_{12}, \ldots, p_{NN}\right\}
    \end{equation}

In addition, we use the average pooling of all clause representations of document to initialize the document node:
    \begin{equation}
        H^{(0)}_D = Avgpool(H) \in \mathbb{R}^d
    \end{equation}

After that, we apply the RGCN on our graph. Given a node $u$, it is defined as:
    \begin{gather}
        s^{(l)}_u = W^{(l)}_s h^{(l)}_u + b^{(l)}_s \\
        t^{(l+1)}_u = s^{(l)}_u + \sum_{r \in \mathcal{R} } \sum_{v \in \mathcal{N}_r (u)} \frac{1}{|\mathcal{N}_r (u)|} W^{(l)}_r h^{(l)}_v  + b^{(l)}_r \\
        h^{(l+1)}_u = ReLU \left( t^{(l+1)}_u \right)
    \end{gather}
where $l$ is the $l$-th layer of RGCN, $\mathcal{R}$ are different types of edges, $W_s^{(l)} \in \mathbb{R}^{d \times d}$, $b^{(l)}_s \in \mathbb{R}^d$, $W^{(l)}_r \in \mathbb{R}^{d \times d}$ and $b^{(l)}_r \in \mathbb{R}^d$ are learnable parameters, $\mathcal{N}_r (u)$ is the neighbours for node $u$ connected with the edge of type $r$, and $ReLU$ is the ReLU activation function.

Finally, we select the last layer as the final representation of all nodes after convolutional operation of $\theta$ layers:
\begin{equation}
    E = H^{(\theta)}_E, \, C = H^{(\theta)}_C, \, P = H^{(\theta)}_P
\end{equation}

\subsection{Classification}
After getting all the representations of nodes, we use a simple MLP to obtain the predictions of emotion-cause pairs:
    \begin{equation}
        \hat{y}^p_{ij} = \sigma \left ( MLP\left( \left[ P_{ij}, E_i, C_j \right] \right) \right)
    \end{equation}
where MLP includes two full-connected layers and a ReLU activation function between them, $\sigma$ is the sigmoid activation function.

Correspondingly, the binary cross entropy loss is utilized as the loss of ECPE:
    \begin{equation}
        \mathcal{L}_p = -\sum_i^N \sum_j^N y^p_{ij} \log(\hat{y}^p_{ij})
    \end{equation}
where $y^p_{ij}$ is the ground truth label.

Following the settings in \citet{wei-etal-2020-effective}, we set two auxiliary tasks which are emotion clauses extraction and cause clauses extraction in order to make the clause nodes learn the key contextual information about emotion or cause in the clauses. We compute the probability as follows:
    \begin{gather}
        \hat{y}^e_i = \sigma \left ( W_e E_i + b_e \right ) \\
        \hat{y}^c_j = \sigma \left ( W_c C_j + b_c \right )
    \end{gather}
where $\hat{y}^e_i$ and $\hat{y}^c_j$ are the probability of emotion and cause clauses separately, $\sigma$ is the sigmoid activation function, $W_e \in \mathbb{R}^{1 \times d}$, $W_c \in \mathbb{R}^{1 \times d}$, $b_e \in \mathbb{R}$ and $b_c \in \mathbb{R}$ are learnable parameters.

Similarly, they have the corresponding loss:
    \begin{gather}
        \mathcal{L}_e = -\sum_i^N y^e_{i} \log(\hat{y}^e_{i}) \\
        \mathcal{L}_c = -\sum_j^N y^c_{j} \log(\hat{y}^c_{j})
    \end{gather}
where $y^e_{i}$ and $y^c_{j}$ are the ground truth labels.

\begin{table}[t]
\centering
\resizebox{\columnwidth}{!}{
\begin{tabular}{lcc}
\hline
\textbf{Item}                            & \textbf{Quantity} & \textbf{Percentage(\%)} \\ \hline
\# of documents                                           & 1,945                              & 100      \\
- w/ 1 pair                               & 1,746                              & 89.77    \\
- w/ 2 pairs                              & 177                                & 9.10     \\
- w/ $\ge 3$ pairs                       & 22                                 & 1.13     \\ \hline
\# of pairs                                 & 2167                               & 100      \\
- w/ 0 relative position         & 511                                & 23.58    \\
- w/ 1 relative position         & 1342                               & 61.93    \\
- w/ 2 relative position         & 224                                & 10.34    \\
- w/ $\ge 3$ relative position & 90                                 & 4.15     \\ \hline
Avg. \# of clauses per document                       & 14.77                              &            \\
Max. \# of clauses per document                       & 73                                 &          \\ \hline
\end{tabular}
}
\caption{The detail of the Chinese corpus.}
\label{tab:dataset}
\end{table}

\begin{table*}[t]
\centering
\resizebox{2 \columnwidth}{!}{

\begin{tabular}{llllllllll}
\hline
\multicolumn{1}{l|}{\multirow{2}{*}{Approach}}                                                                                                                                                                                                           
                                      & \multicolumn{3}{c|}{\textbf{Emotion-Cause Pair Extraction}}                                                                                & \multicolumn{3}{c|}{\textbf{Emotion Clause Extraction}}                                                                                    & \multicolumn{3}{c}{\textbf{Cause Clause Extraction}}                                                                  \\ \cline{2-10} 
\multicolumn{1}{l|}{}                                      & $P$                                & $R$                                & \multicolumn{1}{l|}{$F1$}                               & $P$                                & $R$                                & \multicolumn{1}{l|}{$F1$}                               & $P$                                & $R$                                & $F1$                               \\ \hline
\multicolumn{1}{l|}{ECPE-2D}      & 72.92                              & 65.44                              & \multicolumn{1}{l|}{68.89}                              & 86.27                              & \textbf{92.21}    & \multicolumn{1}{l|}{\underline{89.10}}                              & 73.36                              & 69.34                              & 71.23                              \\
\multicolumn{1}{l|}{TransECPE}                             & \underline{77.08} & 65.32                              & \multicolumn{1}{l|}{70.72}                              & 88.79                              & 83.15                              & \multicolumn{1}{l|}{85.88}                              & 78.74                              & 66.89                              & 72.33                              \\
\multicolumn{1}{l|}{PairGCN}                               & 76.92                              & 67.91                              & \multicolumn{1}{l|}{72.02}                              & 88.57                              & 79.58                              & \multicolumn{1}{l|}{83.75}                              & 79.07    & 68.28                              & 73.75                              \\
\multicolumn{1}{l|}{UTOS}                                  & 73.89                              & 70.62                              & \multicolumn{1}{l|}{72.03}                              & 88.15                              & 83.21                              & \multicolumn{1}{l|}{85.56}                              & 76.71                              & 73.20                              & 74.71                              \\
\multicolumn{1}{l|}{MTST-ECPE$\diamond$}                   & 75.78                              & 70.51                              & \multicolumn{1}{l|}{72.91}                              & 85.83                              & 80.94                              & \multicolumn{1}{l|}{83.21}                              & 77.64                              & 72.36                              & 74.77                              \\
\multicolumn{1}{l|}{RankCP}                                & 71.19                              & \textbf{76.30}    & \multicolumn{1}{l|}{73.60}                              & \textbf{91.23}    & 89.99                              & \multicolumn{1}{l|}{\textbf{90.57}}    & 74.61                              & \underline{77.88} & 76.15                              \\
\multicolumn{1}{l|}{ECPE-MLL\dag}           & 77.00                              & 72.35                              & \multicolumn{1}{l|}{\underline{74.52}} & 86.08                              & \underline{91.91} & \multicolumn{1}{l|}{88.86} & 73.82                              & \textbf{79.12}    & \underline{76.30} \\ \hline
\multicolumn{1}{l|}{PBJE}                                  & \textbf{79.22}    & \underline{73.84} & \multicolumn{1}{l|}{\textbf{76.37*}}    & \underline{90.77} & 86.91                              & \multicolumn{1}{l|}{88.76}                              & \textbf{81.79}    & 76.09                              & \textbf{78.78}    \\ \hline
\end{tabular}
}
\caption{The results comparison with baselines on the ECPE corpus for Emotion-Cause Pair Extraction and the two sub-tasks: Emotion clause Extraction and Cause clause Extraction. We introduce these baselines in Appendix~\ref{sec:appendix comparative approach}. The best performance is in \textbf{bold} and the second best performance is \underline{underlined}. Approach with $\dag$ is previous state-of-the-art method. Approach with $\diamond$ is based on our implementation. * denotes $p < 0.0005$ for a two-tailed t-test against the RankCP.}
\label{tab:results}
\end{table*}

\subsection{Training Object}
We train PBJE by jointly optimizing the three sub-tasks. The total training object is defined as follow:
    \begin{equation}
        \mathcal{L} = \alpha \mathcal{L}_p + \beta \mathcal{L}_e + \gamma  \mathcal{L}_c
    \end{equation}
where $\alpha$, $\beta$ and $\gamma$ are hyper-parameters.

% During inference, since each document has one pair at least, we first extract the pair with the biggest probability $\hat{y}^p_{ij} (1 \leq i, j \leq N)$ as the pair model predict. In addition, for other pairs, we set a threshold $\phi$ to decide if a pair is extracted by model:
%     \begin{equation}
%         {c}^p_{ij} = 
%         \begin{cases}
%             1, &if \quad \hat{y}^p_{ij} > \phi
%             & 0, & otherwise
%         \end{cases}
%     \end{equation}
% where $\phi$ is a hyper-parameter, and ${c}^p_{ij} = 1$ means $(c_i, c_j)$ is a candidate pair.

\section{Experiments}
Extensive experiments are conducted to verify the effectiveness of the PBJE.

\subsection{Dataset and Evaluation Metrics}

We use the Chinese benchmark dataset released by \citet{xia-ding-2019-emotion}, which is constructed from the SINA city news. Table~\ref{tab:dataset} shows the detail of the dataset. Following \citet{xia-ding-2019-emotion}, we use the 10-fold cross-validation as the data split strategy and the precision $P$, recall $R$ and F-score $F1$ as evaluation metrics on three tasks: Emotion-Cause Pair Extraction, Emotion clause Extraction and Cause clause Extraction. We run 10 times and report the average results.

\subsection{Implementation Details}

We implement PBJE based on Transformers\footnote{\url{https://github.com/huggingface/transformers}} \citep{wolf-etal-2020-transformers}, and use the default parameters in BERT-base-Chinese, setting the hidden size $d$ to 768. Additionally, the hyperparameters $\lambda$ and $\theta$ are set to 3 and 1, respectively. The $\alpha$, $\beta$ and $\gamma$ are all set to $1$. We train PBJE through AdamW \citep{loshchilov2018decoupled} optimizer and the learning rate is 2e-5. Meanwhile, we add dropout\citep{10.5555/2627435.2670313} with a rate of 0.2 to avoid over-fitting. Finally, we set the mini-batch to 4 and the training epoch to 35. The experiments are run on the PyTorch-1.9.0 platform and Ubuntu 18.04 using the Intel(R) Core(TM) i7-8700K CPU, 64GB RAM and NVIDIA GeForce RTX 2080 Ti 11GB GPU.

\begin{table*}[t]
\centering
\resizebox{2.0 \columnwidth}{!}{
\begin{tabular}{p{3.9cm}|p{1.1cm}p{1.1cm}p{1.1cm}|p{1.1cm}p{1.1cm}p{1.1cm}|p{1.1cm}p{1.1cm}p{1.1cm}}
\hline
\multirow{2}{*}{Approach} & \multicolumn{3}{c|}{\textbf{Emotion-Cause Pair Extraction}} & \multicolumn{3}{c|}{\textbf{Emotion Clause Extraction}} & \multicolumn{3}{c}{\textbf{Cause Clause Extraction}} \\ 
\cline{2-10}
&$P$ &$R$ &$F1$ &$P$ &$R$ &$F1$ &$P$ &$R$ &$F1$\\
\hline
PBJE &\textbf{79.22} &\textbf{73.84} &\textbf{76.37} &\underline{90.77} &86.91 &88.76 &\textbf{81.79} &\textbf{76.09} &\textbf{78.78}\\
- w/o Clause-Clause Edge &77.81 &\underline{73.36} &\underline{75.45} &90.76 &\underline{87.64} &\textbf{89.14} &80.07 &75.3 &\underline{77.54} \\
- w/o Clause-Pair Edge & \underline{78.14} & 72.62 & 75.21 & 90.76 & 86.74 & 88.66 & \underline{80.15} & 74.51 & 77.16 \\
- w/o Pair Node &76.92 &72.37 &74.54 &89.83 &86.62 &88.18 &79.50 &74.81 &77.05 \\
- w/o PG &78.02 &72.13 &74.93 &\textbf{91.22} &86.73 &\underline{88.89} &80.07 &74.00 &76.89 \\
- w/o Pair Node \& PG &74.49 &73.24 &73.76 &89.93 &\textbf{87.83} &88.82 &78.94 &\underline{75.63} &77.18 \\
% - w/o Doc. Node &\underline{78.24} &71.77 &74.83 &\textbf{91.99} &86.84 &\textbf{89.32} &\underline{80.64} &74.07 &77.18 \\
\hline
\end{tabular}
}
\caption{The results of ablation study on the benchmark corpus for emotion-cause pair extraction and the two sub-tasks. The best performance is in \textbf{bold} and the second best performance is \underline{underlined}.}
\label{tab:ablation}
\end{table*}

\subsection{Overall Results}

Table~\ref{tab:results} shows the results of the Emotion-Cause Pair Extraction (ECPE) task and two sub-tasks: Emotion clause Extraction (EE) and Cause clause Extraction (CE). PBJE shows an apparent advantage over previous work, especially on the main task ECPE and auxiliary task CE. 
% Specifically, PBJE separately obtains $1.85\%$ and $2.77\%$ $F1$ improvements on ECPE compared with the previous best methods ECPE-MLL and RankCP. 
We argue that the joint encoding manner plays an important role in PBJE, making the interaction bidirectional and balancing the information obtained by pairs and clauses. It is worth noting that PBJE shows a significant improvement on CE while demonstrating a similar performance on EE compared with ECEP-MLL, which means PBJE can balance the EE and CE. Specifically, RankCP has a huge improvement on EE with applying the sentiment lexicon to PBJE. However, it achieves poor performance on CE, leading to a sharp drop on ECPE. Similarly, ECPE-2D encounters the imbalance problem compared with PBJE. It obtains the second best result on EE, but the worst result on CE. In most cases, EE is more difficult to cope with \citep{xia-ding-2019-emotion}. Because the expression about cause often contains multiple words, and thus requires the models to understand the text. On the contrary, the expression about emotion only contains a single keyword (e.g., angry, as shown in Figure~\ref{fig:example}).
We argue that the balance benefits from modeling two types of clauses efficiently. And further, this balance helps PBJE improve performance on ECPE.

\subsection{Ablation Study}
Ablation studies are conducted to verify the effectiveness of the Pair Generator (PG) and different relationship edges and nodes in our graph. Table~\ref{tab:ablation} shows the results of the ablation studies.

\noindent\textbf{w/o Clause-Clause Edge} \quad We use one type of edge to replace the Clause(Emotion)-Clause(Emotion) Edge and the Clause(Cause)-Clause(Cause) Edge.
% It means the model does not distinguish the emotion clause nodes and the cause clause nodes, but the two types of Clause-Pair Edge still extract different information from these nodes.
Without these two edges, the performance of our model dramatically drops on CE, further leading to the drop on ECPE. It breaks the balance between EE and CE, meaning the model tends to focus on EE but neglect CE, since EE is the earliest task among these three tasks.
% The contextual information in emotion clauses is different from those in cause clauses. Therefore, using the same representations for prediction in EE and CE will blur their features and lead to a drop in results.

\noindent\textbf{w/o Clause-Pair Edge} \quad We remove the Clause(Emotion)-Pair Edge and Clause(Cause)-Pair, and use another edge to replace them.
% It means the pairs interact with the emotion clauses and cause clauses in the same way.
The performance on ECPE is even worse than w/o Clause-Clause Edge. The pairs separately extract the emotions from emotion clauses and the reason from cause clauses. Without this difference, the model can hardly extracts information efficiently for the causal relationship.

\noindent\textbf{w/o Pair Node} \quad We remove the pair nodes and separately model the emotion and cause clauses using the RGCN. The pairs from PG are utilized to replace the pairs after RGCN.
% , and they are concatenated with the clauses after RGCN for prediction. 
In this way, it is a typical sequential encoding method.
% The two types of clauses can not interact, and the pairs can not interact together with the clauses to learn the causal relationship. 
Although the PG can still provide some information between emotion and cause clauses, it generates the second worst result on F1. The result shows the importance of joint encoding manner and the causal relationship.

\noindent\textbf{w/o PG} \quad Meanwhile, we remove the PG and use another relative position embedding to replace the representations of pairs, which means the pairs with the same relative positions will have the same initial representations in the RGCN and do not contain any clause information. Without the PG, the performance
% also drops on ECPE, but it 
is still better than w/o Pair Node. Despite of absence of clause information, the pairs can learn clause features and causal relationship by the Clause-Pair Edge, which also indicates that the causal relationship is crucial to the modeling of pairs emerging from joint encoding manner. 
% Therefore, compared with the model which can not learn the causal relationship without pair nodes, the model without PG achieves better performance on ECPE.

\noindent\textbf{w/o Pair Node \& PG} \quad Moreover, we remove the pair nodes and PG together, similar to the methods in previous work which only encode the clauses for prediction. The $F1$ on ECPE sharply dropped by 2.61$\%$ and it is the worst model in our experiments. We argue it is caused by the ignorance of pair modeling and the causal relationship in pairs.

% \noindent\textbf{w/o Doc. Node} \quad Finally, we remove the document node in the RGCN. The drop in performance mainly occurs in ECPE and CE. We believe that the global information of documents (e.g., topics) is beneficial for the ECPE. We will explore it in detail in Appendix~\ref{sec:doc}.
In addition, we perform some fine-grained experiments to verify the effect of document node in Appendix~\ref{sec:doc}. We believe the information of documents (e.g., topics) is beneficial for the ECPE.

\begin{table}[t!]
\centering
\resizebox{0.8 \columnwidth}{!}{
\begin{tabular}{ccccc}
\hline
 \#Pairs                                                                       & Approach & $P$   & $R$   & $F1$  \\ \cline{1-5} 
                                                                                  \multirow{2}{*}{1 per doc.}                                                   & PBJE     & \textbf{78.44} & 80.00 & \textbf{79.21}  \\
                                                                                                                                                                & RankCP   & 72.03 & \textbf{81.23} & 76.33 \\ \cline{1-5} 
                                                                                  \multirow{2}{*}{\begin{tabular}[c]{@{}c@{}}2 or more\\ per doc.\end{tabular}} & PBJE     & \textbf{83.98} & 45.29 & \textbf{58.84} \\
                                                                                                                                                                & RankCP   & 67.72 & \textbf{51.46} & 58.02 \\ \hline
\end{tabular}
}
\caption{The results of ECPE for documents with different numbers of pairs.}
\label{tab:single_multi_pair}
\end{table}

\subsection{The Effect of Joint Encoding Manner}
To verify the effect of taking into account the joint encoding manner in ECPE, we further conduct some experiments in special cases.
% Due to the great improvement on multi pairs with sentiment lexicon, we separately implement RankCP and PBJE with or without sentiment lexicon for a fair comparison.

We first compare the results in two situations: documents with one ground truth pair and documents with two or more ground truth pairs. The results are shown in Table~\ref{tab:single_multi_pair}. PBJE shows clear superiority in both situations. In the documents with a single pair, PBJE demonstrates a significant improvement on F1 on ECPE, because it avoids the problem caused by sequential encoding. Specifically, in sequential encoding, if the clause encoders extract wrong emotion or cause clauses, the pair encoder is prone to easily group the wrong clauses and extract them as a pair without the bidirectional interaction between clauses and pairs. 
Apart from that, PBJE also shows a competitive improvement on the document with multiple pairs, which means PBJE can take into account more about the causal relationship
% . When a document has one pair, the model just needs simply group the emotion clause and cause clause extracted by the clause encoders. However, when a document has multi pairs, the model needs to 
and handle the situation mentioned in Figure~\ref{fig:example}. 
% The improvement on precision on multi pairs situation demonstrates that PBJE can reduce the predictions without causal relationship, and finally, it reflects in the improvement of F1.
% We argue the inter-relationship among emotion clauses, cause clauses, and pairs plays an important role in PBJE, which does not exist in the sequential encoding methods. The inter-relationship makes the model focus more on causal relationship.

\begin{table}[t]
\centering
\resizebox{0.8 \columnwidth}{!}{
\begin{tabular}{ccccc}
\hline
\multirow{2}{*}{\begin{tabular}[c]{@{}c@{}}Relative\\ Position\end{tabular}} & \multirow{2}{*}{Approach} & \multirow{2}{*}{P} & \multirow{2}{*}{R} & \multirow{2}{*}{F} \\
                                                                                 \\ \hline
  \multirow{2}{*}{$\leq 1$}                                               & PBJE                      & \textbf{80.69}              & 81.26              & \textbf{80.97}              \\
                                                                                 &                                                                               RankCP                    & 77.45              & \textbf{83.38}              & 80.30               \\ \cline{1-5} 
                                                                                 \multirow{2}{*}{$\ge 2$}                                            & PBJE                      & \textbf{58.55}              & 28.43               & \textbf{38.28}              \\
                                                                                 &                                                                               RankCP                    & 31.60               & \textbf{32.91}             & 32.24              \\ \hline
\end{tabular}
}
\caption{The results of ECPE for pairs of different relative positions.}
\label{tab:relative_distance}
\end{table}

\begin{table}[t]
\centering
\resizebox{\columnwidth}{!}{
\begin{tabular}{llll}
\hline
\multicolumn{4}{l}{\begin{tabular}[c]{@{}l@{}}...It's time for Chinese New Year.($c_4$) The creditor removed all \\ his family's grain.($c_5$) Other families are celebrating the New \\ Year happily.($c_6$) But his family \textcolor{Blue}{even did not have money for} \\ \textcolor{Blue}{meat}.($c_7$) His daughter and wife \textcolor{Red}{sorrowed}.($c_8$)...\end{tabular}} \\ \hline
PBJE                                                           & \multicolumn{1}{l|}{\textcolor{Green}{{[}$c_8$,$c_7${]}}}                                                         & RankCP                                                        &  \textcolor{Red}{{[}$c_6$,$c_5${]}},\textcolor{Red}{{[}$c_6$,$c_7${]}},\textcolor{Green}{{[}$c_8$,$c_7${]}}                                                        \\ \hline
Ground Truth                                                         & \multicolumn{3}{l}{{[}$c_8$,$c_7${]}}                                                                                                                                                                         \\ \hline
\end{tabular}

}
\caption{An example predicted by PBJE and RankCP. The words in \textcolor{Red}{red} are the emotion keywords, and the words in \textcolor{Blue}{blue} are the cause keywords. The pairs in \textcolor{Green}{green} are the correct prediction, and the pairs in \textcolor{Red}{red} are incorrect. We translate it from Chinese into English for ease of reading.}
\label{tab:case}
\end{table}

In addition, we compare the results in another two situations, which are pairs $(c_i,c_j)$ with relative position less or equal than 1 $(|i-j| \leq 1)$ and greater or equal than 2 $(|i-j| \ge 2)$. The results are shown in Table~\ref{tab:relative_distance}. Because of the linguistic expression habits, the benchmark dataset has a strong position bias \citep{ding2020experimental, Xia2019RTHNAR, Ding_He_Zhang_Xia_2019}. Most of the pairs in the dataset have a relative position within 1 (about 85\%). 
The models can easily achieve a good performance with enough training data in this situation.
% If the models efficiently utilize relative position embedding, they can easily achieve high performance by correctly extracting the pairs of relative position within 1. 
However, extracting the more difficult pairs are ignored, and these pairs are still prevalent in the real world. 
Apart from getting a significant improvement in the relative position within 1, PBJE surpasses RankCP in the relative position greater or equal to 2. It shows that PBJE can handle the more complex situation.
% although PBJE is poorly trained in this situation for lack of training data whose relative position is greater than 1. 
We argue that the causal relationship in joint encoding manner contributes to PBJE the most. Since PBJE needs to consider the causal relationship among multiple clauses and filter irrelevant clauses, when the relative position is greater than 2 and it can not supply enough information.

\subsection{Case Study}
\begin{figure}[t]
    \centering
    \includegraphics[width=\linewidth]{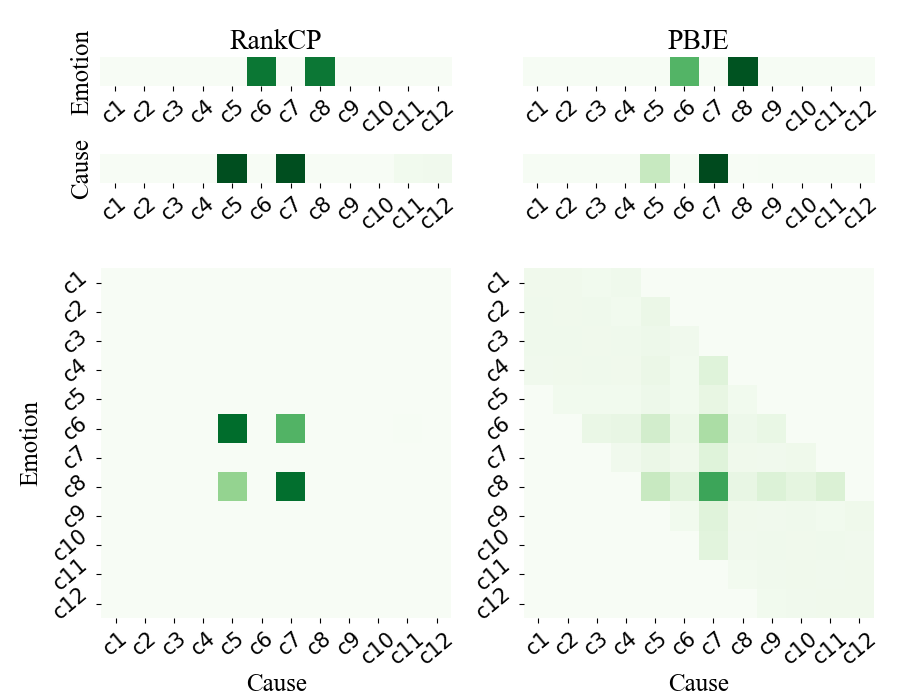}
    \caption{Visualization of the confidence of each prediction in RankCP and PBJE. The deeper color means the higher confidence.}
    \label{fig:heatmap}
\end{figure}

We analyze an example selected from the benchmark corpus to demonstrate the effectiveness of joint encoding manner and considering the causal relationship in PBJE, which is shown in Table~\ref{tab:case}. In addition, we visualize the prediction results in Figure~\ref{fig:heatmap}.
% The ground truths and the predicted results of PBJE and RankCP are shown in Table~\ref{tab:case}.
% We choose the RankCP to compare with PBJE because it is more representative.

In this example, RankCP and PBJE both extract the 
% correct clause $c_8$ and a 
wrong clause $c_6$ in EE, which
% . The incorrect extraction 
might be attributable to the word "happy" in the text. Although $c_6$ expresses happiness for the Chinese New Year, there is no corresponding cause clause. We do not define it as an emotion clause. Moreover, in CE, RankCP extracts a wrong cause clause $c_5$, but PBJE does not (with about 0.25 confidence). The difference shows the clause encoding capability of PBJE. It is worth noting that, in ECPE, the sequential encoding method RankCP tends to easily couple the clause extracted in EE and CE, and it can not correct the errors in EE and CE, resulting in extracting another two wrong pairs. On the contrary, PBJE can avoid them and find out the most likely pair by the joint encoding manner and the balance of information flow between pairs and clauses.

In addition, we analyze more examples in Appendix~\ref{sec:appendix case study}. We can find that RankCP tends to extract as many candidate pairs as possible. But there could be many wrong pairs. 
On the contrary, PBJE tries to extract the correct pairs directly. It can explain why PBJE performs better on precision and worse on recall compared with RankCP.

\section{Conclusion}

In this paper, we propose a novel \textbf{P}air-\textbf{B}ased \textbf{J}oint \textbf{E}ncoding (\textbf{PBJE}) network, which encodes the pairs and clauses features simultaneously. It can balance the inter-task feature interaction compared with sequential encoding and model the causal relationship between emotion clauses and corresponding cause clauses by pairs.
Furthermore, it can avoid the wrong predictions in previous tasks in the multi-task learning. 
From a multi-relational perspective, we propose a Relational Graph Convolutional Network (RGCN) framework to capture the relationship among emotion clauses, cause clauses, pairs, and document, including four types of node and five types of edge. The experiments on the Chinese benchmark corpus show that PBJE achieves state-of-the-art performance.
% And the extensive experiments show the effectiveness of each component in the graph and the advantages of our model in special cases.

% In our future work, we will further try to handle the multiplex relationship among all nodes and overcome the over-smoothing problem in the RGCN.

\section*{Limitations}
Following the previous work, we implement PBJE setting the hyperparameter $\lambda$ to 3. It means PBJE can only extract the pairs with a relative distance less than or equal to 3 ($i - j \le 3$). However, the maximum relative distance of pairs in the dataset is 12. Therefore, no matter how good PBJE is, it can not extract all pairs. There are some methods to solve this problem. For example, we can set $\lambda$ to 12. Furthermore, we can enumerate all possible pairs without limiting relative distance ($\lambda = +\infty$) to satisfy a larger relative distance that may occur in application. Nonetheless, these two methods will slightly affect the performance of PBJE. Because when we increase $\lambda$, there will be more negative samples in the dataset, which exacerbates the problem of data imbalance. On the other hand, because of the language expression, the emotion clauses and cause clauses co-occur most of the time. Setting a large $\lambda$ is unnecessary. Therefore, the tradeoff between relative distance and performance is what we need to explore in the future. Additionally, a new method which is free from the influence of the relative distance is more desirable.

\section*{Acknowledgements}
We thank the anonymous reviewers for their helpful feedbacks. The work described in this paper was partially funded by the National Natural Science Foundation of China (Grant Nos. 62272173, 61872148), the Natural Science Foundation of Guangdong Province (Grant Nos. 2022A1515010179, 2019A1515010768).

\bibliography{custom}
\bibliographystyle{acl_natbib}

\appendix

\section{Comparative Approaches}
\label{sec:appendix comparative approach}

We compare PBJE with the following methods, which use the pre-trained BERT as encoder:

\begin{itemize}
    \item \textbf{ECPE-2D} \citep{ding-etal-2020-ecpe}: This method uses the 2D representation to construct a pairs matrix and utilizes the 2D transformer module to interact with other pairs for prediction.
    \item \textbf{TransECPE} \citep{fan-etal-2020-transition}: It is a transition-based method which transforms the task into a procedure of parsing-like directed graph construction.
    \item \textbf{RankCP} \citep{wei-etal-2020-effective}: This method tackles emotion-cause pair extraction from a ranking perspective, which ranks pairs in a document and proposes a one-step neural approach to extract.
    \item \textbf{PairGCN} \citep{chen-etal-2020-end}: This method constructs a graph using the pair nodes and a Pair Graph Convolutional Network to model the dependency relations among candidate pairs.
    \item \textbf{ECPE-MLL} \citep{ding-etal-2020-end}: It is the current state-of-the-art method, which employs two joint frameworks, including the emotion-pivot cause extraction and cause-pivoted emotion extraction with sliding window strategy.
    \item \textbf{UTOS} \citep{9511845}: It solves this task using sequence labeling, which allows to extract pairs through one pass and addresses the error propagation problem.
    \item \textbf{MTST-ECPE} \citep{9457144}: This method uses a multi-task sequence tagging framework with refining the tag distribution.
\end{itemize}

\begin{table}[t!]
    \centering
    \resizebox{\columnwidth}{!}{
    \begin{tabular}{lclccc}
    \hline
    \begin{tabular}[c]{@{}l@{}}\#Clauses \\ per Doc.\end{tabular} & \% in Corpus           & Approach      & $P$   & $R$   & $F1$  \\ \hline
    \multirow{3}{*}{$< 14$}                                 & \multirow{3}{*}{45.71} & PBJE          & \textbf{81.26} & 75.97 & \textbf{78.53} \\
                                                                  &                        & - w/o Doc. Node & 78.90 & 72.71 & 75.68 \\
                                                                  &                        & RankCP        & 69.82 & \textbf{77.49} & 73.46 \\ \hline
    \multirow{3}{*}{$\ge 14$}                                   & \multirow{3}{*}{54.29} & PBJE          & 77.11 & 72.02 & \textbf{74.48} \\
                                                                  &                        & - w/o Doc. Node & \textbf{77.49} & 70.9 & 74.05 \\
                                                                  &                        & RankCP        & 72.00 & \textbf{75.28} & 73.60 \\ \hline
    \end{tabular}
    }
    \caption{The result of ECPE for documents with different numbers of clauses.}
    \label{tab:doclen}
\end{table}

\begin{table*}[t]
\centering
\resizebox{2 \columnwidth}{!}{
\begin{tabular}{llccc}
\hline
\multirow{2}{*}{ID} & \multirow{2}{*}{Examples}                                                                                                                                                                                                                                                                                                                                                                            & \multicolumn{2}{c}{Predicted Pairs}                                                                                                                 & \multirow{2}{*}{\begin{tabular}[c]{@{}c@{}}Ground\\ Truths\end{tabular}} \\ \cline{3-4}
                    &                                                                                                                                                                                                                                                                                                                                                                                                     & PBJE                                                              & RankCP                                                                          &                                                                          \\ \hline
1                   & \begin{tabular}[c]{@{}l@{}}Mr. Zhang, the boss of the repair shop where he used to be, said.($c_{1}$) Mr. Wang \\ was very \textcolor{Red}{surprised} when he \textcolor{Blue}{came to apply}.($c_{2}$) He was the most educated worker \\ in the garage at the time.($c_{3}$) Every time he fixed the cars, he put his heart into \\ it.($c_{4}$) Once he was \textcolor{Blue}{ill}, he still \textcolor{Blue}{insisted on repairing the car}.($c_{5}$) It made Mr. Zhang \\ \textcolor{Red}{so moved}.($c_{6}$) ...
\end{tabular}                                                                        & \begin{tabular}[c]{@{}c@{}}\textcolor{Green}{{[}$c_{2}$, $c_{2}${]}}\\ \textcolor{Green}{{[}$c_{6}$, $c_{5}${]}}\end{tabular} & \begin{tabular}[c]{@{}c@{}}\textcolor{Green}{{[}$c_{2}$, $c_{2}${]}}\\ \textcolor{Red}{{[}$c_{2}$, $c_{5}${]}}\\ \textcolor{Green}{{[}$c_{6}$, $c_{5}${]}}\end{tabular} & \begin{tabular}[c]{@{}c@{}}{[}$c_{2}$, $c_{2}${]}\\ {[}$c_{6}$, $c_{5}${]}\end{tabular}        \\ \hline
2                   & \begin{tabular}[c]{@{}l@{}}... She and her family are very healthy($c_2$) So they can  continue to donate blood \\ to contribute to the society($c_3$) She used to \textcolor{Red}{worry} about \textcolor{Blue}{limiting the age of blood} \\ \textcolor{Blue}{donation to 55 years old}($c_4$) She is used to donating blood now($c_5$) If she \textcolor{Blue}{can't} \\\textcolor{Blue}{continue donating blood because of her age}, she will be very \textcolor{Red}{disappointed}($c_6$) ...\end{tabular} & \begin{tabular}[c]{@{}c@{}}\textcolor{Green}{{[}$c_4$, $c_4${]}}\\ \textcolor{Green}{{[}$c_6$, $c_6${]}}\end{tabular}     & \begin{tabular}[c]{@{}c@{}}\textcolor{Green}{{[}$c_4$, $c_4${]}}\\ \textcolor{Red}{{[}$c_4$, $c_6${]}}\\ \textcolor{Green}{{[}$c_6$, $c_6${]}}\end{tabular}       & \begin{tabular}[c]{@{}c@{}}{[}$c_4$, $c_4${]}\\ {[}$c_6$, $c_6${]}\end{tabular}            \\ \hline
\end{tabular}
}
\caption{Examples predicted by PBJE and RankCP. The words in \textcolor{Red}{red} are the emotion keywords, and the words in \textcolor{Blue}{blue} are the cause keywords. The pairs in \textcolor{Green}{green} are the correct prediction, and the pairs in \textcolor{Red}{red} are incorrect. We translate them from Chinese into English for ease of reading.}
\label{app:case}
\end{table*}

\section{The Effect of Document Node}
\label{sec:doc}
To verify the effect of document node, some extensive experiments are conducted in different lengths of a document, according to the average number of clauses per document $14.77$ and the median $14$. 
% Besides, we consider the extreme case in which a document contains 20 clauses or more.

As shown in Table~\ref{tab:doclen}, the document node can help PBJE to improve the performance on ECPE in both short ($<14$) and long ($\ge 14$) documents. Since the emotion clauses and cause clauses make up a small proportion of the total clauses in each document, even in the short documents. Most of the documents only have 1 pair. Therefore, the fully connected graphs of emotion and cause clauses contain lots of noise. It makes each emotion and cause clause node can hardly learn the effective and enough contextual information. In this situation, the document node can filter the invalid information and integrate them into global information, then transmits them to other nodes through the Document-Others Edge. However, the improvement in short documents on ECPE is much more than in long documents with the help of document node.
% And in the extremely long documents which contain 20 clauses or more, the improvement is also small. 
Because when the document is long, there are too many features of clauses for average pooling. This results in a lower effective information density in long documents than in short documents. Further, it makes the representations of document uncharacteristic and contain noise. But the performance does not drop in this situation since the Document-Others Edge can selectively transmit information through learning ability.

\section{Additional Case Study}
\label{sec:appendix case study}
To further demonstrate the importance of considering the causal relationship in RGCN, we analyze another two examples selected from the benchmark corpus. We show them in Table~\ref{app:case}

For the first example, although RankCP extracts all the correct pairs, it extracts another incorrect pair $(c_{2}, c_{5})$.
The clause $c_{2}$ expresses surprise, and the clause $c_{5}$ expresses the persistence of fixing cars even when he is sick.
Although they are emotion clause and cause clause separately, $c_{2}$ is not the reason to cause $c_{5}$ obviously. By considering the causal relationship, PBJE avoids this situation. 

Next, for the second example, RankCP encountered the same problem as the first example. Further, the emotion clause $c_4$ expresses worry, and $c_6$ expresses disappointment, which are both negative emotions. Moreover, the cause clause $c_4$ describes the same thing with cause clause $c_6$ about the age restrictions on blood donation. Therefore, it is more difficult for models to judge in this situation. Nevertheless, PBJE successfully deals with this situation.

\section{Hyperparameters Discussion}
\begin{figure}[t]
    \centering
    \includegraphics[width=\linewidth]{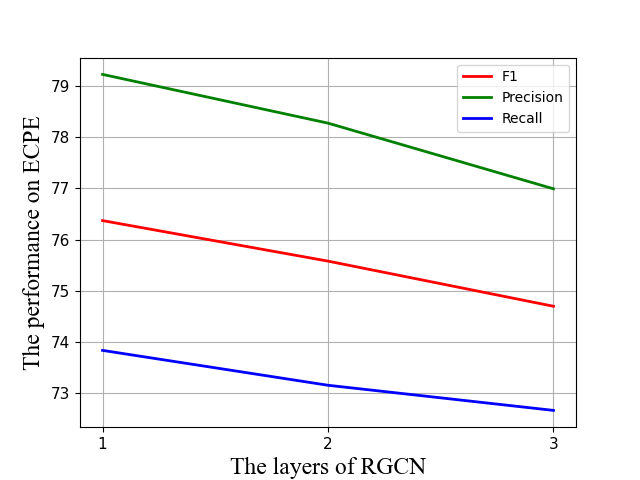}
    \caption{The influence of different $\theta$ on ECPE.}
    \label{fig:hyperparameter}
\end{figure}

As shown in Figure~\ref{fig:hyperparameter}, we examine the effects of different values of $\theta$ on ECPE. We can observe that the performance tends to drop with the increase of the layers of RGCN. We argue that the over-smoothing causes this problem \citep{li2018deeper, ZHOU202057, Chen_Lin_Li_Li_Zhou_Sun_2020}. Specifically, when the $\theta$ is greater than 1, it means that the Relationship Graph Convolutional Network(RGCN) are repeatedly applied. It may mix the features of nodes from different classes and make them indistinguishable, leading to the drop on ECPE. In addition, more layers indicates more learnable parameters, which will result in over-fitting \citep{10.1145/3459637.3482488, rong2020dropedge}.

\end{document}